\documentclass{article}

\usepackage{microtype}
\usepackage{graphicx}
\usepackage{booktabs} 

\usepackage{hyperref}

\usepackage[review]{icml2020}

\usepackage{amssymb}  
\usepackage{dsfont}  
\usepackage{mathtools}  
\usepackage{pbox}  
\newcommand*{\V}[1]{\mathbf{#1}}
\newcommand*{\transpose}[1]{{#1}^\mathsf{T}}

\RequirePackage{algorithm}
\RequirePackage{algorithmic}

\usepackage{amsthm}
\usepackage{caption}
\usepackage{subcaption}
\usepackage{float}
\newcommand*{\floor}[1]{\left \lfloor {#1} \right \rfloor}
\DeclareMathOperator{\expm}{expm1}
\DeclareMathOperator{\logp}{log1p}
\DeclareMathOperator{\sigmoid}{sigmoid}
\newtheorem{theorem}{Theorem} 
\newtheorem{lemma}{Lemma} 
\newtheorem{corollary}{Corollary} 

\icmltitlerunning{A Spike in Performance: Training Hybrid-Spiking Neural Networks with Quantized Activation Functions}

\begin{document}

\twocolumn[
\icmltitle{A Spike in Performance: Training Hybrid-Spiking Neural Networks \\ with Quantized Activation Functions}





\begin{icmlauthorlist}
\icmlauthor{Aaron R. Voelker}{abr}
\icmlauthor{Daniel Rasmussen}{abr}
\icmlauthor{Chris Eliasmith}{abr,uw}
\end{icmlauthorlist}

\icmlaffiliation{abr}{Applied Brain Research Inc.}
\icmlaffiliation{uw}{University of Waterloo, Waterloo, ON, Canada}

\icmlcorrespondingauthor{Aaron R. Voelker}{aaron.voelker@appliedbrainresearch.com}

\icmlkeywords{Hybrid-Spiking Neural Networks, Quantized Activation Functions, Legendre Memory Units, Recurrent Neural Networks, Signal Processing, Nengo}

\vskip 0.3in
]



\printAffiliationsAndNotice{\icmlEqualContribution} 

\begin{abstract}
The machine learning community has become increasingly interested in the energy efficiency of neural networks.
The Spiking Neural Network~(SNN) is a promising approach to energy-efficient computing, since its activation levels are quantized into temporally sparse, one-bit values (i.e.,~``spike'' events), which additionally converts the sum over weight-activity products into a simple addition of weights (one weight for each spike).
However, the goal of maintaining state-of-the-art~(SotA) accuracy when converting a non-spiking network into an SNN has remained an elusive challenge, primarily due to spikes having only a single bit of precision.
Adopting tools from signal processing, we cast neural activation functions as quantizers with temporally-diffused error, and then train networks while smoothly interpolating between the non-spiking and spiking regimes.
We apply this technique to the Legendre Memory Unit~(LMU) to obtain the first known example of a hybrid SNN outperforming SotA recurrent architectures---including the LSTM, GRU, and NRU---in accuracy, while reducing activities to at most 3.74 bits on average with 1.26 significant bits multiplying each weight.
We discuss how these methods can significantly improve the energy efficiency of neural networks.
\end{abstract}

\section{Introduction}

The growing amount of energy consumed by Artificial Neural Networks~(ANNs) has been identified as an important problem in the context of mobile, IoT, and edge applications~\citep{moloney2016embedded, zhang2017hello, mckinstry2018discovering, wang2019energy, tinyml}.
The vast majority of an ANN's time and energy is consumed by the multiply-accumulate~(MAC) operations implementing the weighting of activities between layers~\citep{sze2017efficient}.
Thus, many ANN accelerators focus almost entirely on optimizing MACs~\citep[e.g.][]{ginsburg2017tensor, jouppi2017datacenter}, while practitioners prune~\citep{zhu2017prune, liu2018memory} and quantize~\citep{gupta2015deep, courbariaux2015binaryconnect, mckinstry2018discovering, nayak2019bit} weights to reduce the use and size of MAC arrays.

While these strategies focus on the weight matrix,  the Spiking Neural Network~(SNN) community has taken a different but complementary approach that instead focuses on temporal processing.
The operations of an SNN are \emph{temporally sparsified}, such that an accumulate only occurs whenever a ``spike'' arrives at its destination.
These sparse, one-bit activities (i.e., ``spikes'') not only reduce the volume of data communicated between nodes in the network~\citep{furber2014spinnaker}, but also replace the multipliers in the MAC arrays with adders -- together providing orders of magnitude gains in energy efficiency~\citep{davies2018loihi, blouw2019benchmarking}.


The conventional method of training an SNN is to first train an ANN, replace the activation functions with spiking neurons that have identical firing rates~\citep{hunsberger2015spiking}, and then optionally retrain with spikes on the forward pass and a differentiable proxy on the backward pass~\citep{huh2018gradient, bellec2018long, zhang2019spike}.
However, converting an ANN into an SNN often degrades model accuracy -- especially for recurrent networks.
Thus, multiple hardware groups have started building hybrid architectures that support ANNs, SNNs, and mixtures thereof~\citep{liu2018memory, pei2019towards, tapson2020} -- for instance by supporting event-based activities, fixed-point representations, and a variety of multi-bit coding schemes.
These hybrid platforms present the alluring possibility to trade accuracy for energy in task-dependent ways~\citep{anon2020}.
However, principled methods of leveraging such trade-offs are lacking.

In this work, we propose to our knowledge the first method of training hybrid-spiking networks~(hSNNs) by smoothly interpolating between ANN (i.e.,~32-bit activities) and SNN (i.e.,~1-bit activities) regimes.
The key idea is to interpret spiking neurons as one-bit quantizers that diffuse their quantization error across future time-steps -- similar to \citet{floyd1976adaptive} dithering.
This idea can be readily applied to any activation function at little additional cost, generalizes to quantizers with arbitrary bit-widths (even fractional), provides strong bounds on the quantization errors, and relaxes in the limit to the ideal ANN.

Our methods enable the training procedure to balance the accuracy of ANNs with the energy efficiency of SNNs by evaluating the continuum of networks in between these two extremes.
Furthermore, we show that this method can train hSNNs with superior accuracy to ANNs and SNNs trained via conventional methods.
In a sense, we show that it is useful to think of spiking and non-spiking networks as extremes in a continuum.
As a result, the family of hSNNs captures networks with any mixture of activity quantization throughout the architecture.


\section{Related Work}


Related work has investigated the quantization of activation functions in the context of energy-efficient hardware~\citep[e.g.,][]{jacob2018quantization, mckinstry2018discovering}.
Likewise, \citet{hopkins2019stochastic} consider stochastic rounding and dithering as a means of improving the accuracy of spiking neuron models on low-precision hardware relative to their ideal ordinary differential equations~(ODEs).
These approaches do not account for the quantization errors that accumulate over time, whereas our proposed method keeps them bounded.

Some have viewed spiking neurons as one-bit quantizers, or analog-to-digital~(ADC) converters, including \citet{chklovskii2012neuronal, yoon2016lif, ando2018dither, neckar2018braindrop, yousefzadeh2019conversion, yousefzadeh2019asynchronous}.
But these methods are not generalized to multi-bit or hybrid networks, nor leveraged to interpolate between non-spiking and spiking networks during training.

There also exist other methods that introduce temporal sparsity into ANNs.
One such example is channel gating~\citep{hua2019channel}, whereby the channels in a CNN are dynamically pruned over time.
Another example is dropout~\citep{srivastava2014dropout} -- a form of regularization that randomly drops out activities during training.
The gating mechanisms in both cases are analogous to spiking neurons.

Neurons that can output multi-bit spikes have been considered in the context of packets that bundle together neighbouring spikes~\citep{krithivasan2019dynamic}.
In contrast, this work directly computes the number of spikes in $\mathcal{O}(1)$ time and memory per neuron, and varies the temporal resolution during training to interpolate between non-spiking and spiking and allow optimization across the full set of hSNNs.



Our methods are motivated by some of the recent successes in training SNNs to compete with ANNs on standard machine learning benchmarks~\citep{bellec2018long, stockl2019recognizing, pei2019towards}.
To our knowledge, this work is the first to parameterize the activation function in a manner that places ANNs and SNNs on opposite ends of the same spectrum.
We show that this idea can be used to convert ANNs to SNNs, and to train hSNNs with improved accuracy relative to pure (i.e.,~1-bit) SNNs and energy efficiency relative to pure (i.e.,~32-bit) ANNs.


\begin{figure*}
\begin{center}
\centerline{\includegraphics[width=\textwidth]{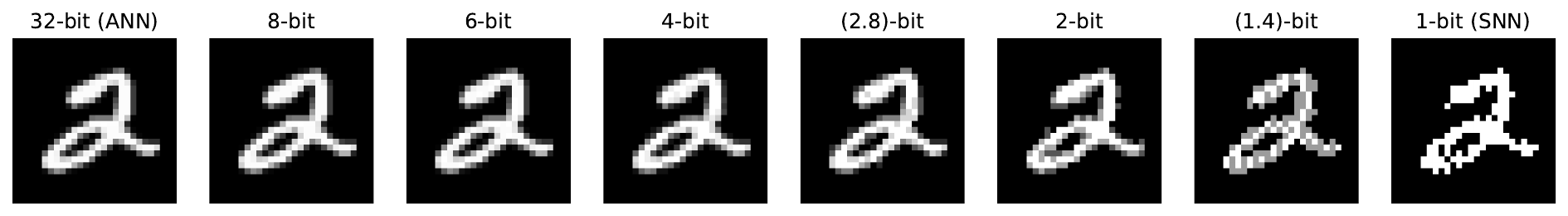}}
\caption{Visualizing the output ($\tilde{a}_t$) of Algorithm~\ref{alg:quantizer} 
given an MNIST digit as input ($x_t$).
The bit-width is varied as $\omega = 2^m - 1$;
$m = 32$ correspond to the activities of a 32-bit ANN, whereas $0 < m \le 1$ correspond to those of an SNN.
}
\label{fig:dither}
\end{center}
\end{figure*}

\section{Methods}
\label{sec:methods}

\subsection{Quantized Activation Functions}

We now formalize our method of quantizing any activation function.
In short, the algorithm quantizes the activity level and then pushes the quantization error onto the next time-step -- analogous to the concept of using error diffusion to dither a one-dimensional time-series~\citep{floyd1976adaptive}.
The outcome is a neuron model that interpolates an arbitrary activation function, $f$, between non-spiking and spiking regimes through choice of the parameter $\omega > 0$, which acts like a time-step.

\subsubsection{Temporally-Diffused Quantizer}

Let $x_t$ be the input to the activation function at a discrete time-step, $t > 0$, such that the ideal output (i.e., with unlimited precision) is $a_t = f(x_t)$.
The algorithm maintains one scalar state-variable across time, $v_t$, 
that tracks the total amount of quantization error that the neuron has accumulated over time.
We recommend initializing $v_0 \sim \mathcal{U}[0, 1)$ independently for each neuron.
The output of the neuron, $\tilde{a}_t$, is determined by Algorithm~\ref{alg:quantizer}.\footnote{%
This algorithm, and related methods, are patent pending technologies~\citep{multibitpatent} of Applied Brain Research Inc.
}

The ideal activation, $f$, may be any conventional nonlinearity (e.g.,~$\tanh$, sigmoid, etc.), or the time-averaged response curve corresponding to a biological neuron model (e.g.,~leaky integrate-and-fire) including those with multiple internal state-variables~\citep{koch2004biophysics}.
Adaptation may also be modelled by including a recurrent connection from $\tilde{a}_t$ to $x_{t+1}$~\citep[][section~5.2.1]{voelker2019}.

To help understand the relationship between this algorithm and spiking neuron models, it is useful to interpret $\tilde{a}_t$ as the number of spikes~($k$) that occur across a window of time normalized by the length of this window~($\omega$).
Then $f(x_t) \times \omega$ represents the expected number of spikes across the window, and $v_t$ tracks progress towards the next spike.

\begin{algorithm}
   \caption{Temporally-Diffused Quantizer ($f$; $\omega$)}
   \label{alg:quantizer}
\begin{algorithmic}
   \STATE {\bfseries Input:} $x_t$
   \STATE {\bfseries State:} $v_t$
   \STATE {\bfseries Output:} $\tilde{a}_t$
   \vspace{0.3em}
   \STATE $s \leftarrow v_{t-1} + f(x_t) \times \omega$
   \STATE $k \leftarrow \left \lfloor s \right \rfloor$
   \STATE $v_t \leftarrow s - k$
   \STATE $\tilde{a}_t \leftarrow k / \omega$
\end{algorithmic}
\end{algorithm}

We note that Algorithm~\ref{alg:quantizer} is equivalent to \citet[][Algorithm~1]{ando2018dither} where $f(x) = \max \left( x \text{,} \, 0 \right)$ is the rectified linear~(ReLU) activation function, and $\omega = 1$.
\citet[][Algorithm~1]{yousefzadeh2019conversion} extend this to represent changes in activation levels, and allow negative spikes.
Still considering the ReLU activation, Algorithm~\ref{alg:quantizer} is again equivalent to the spiking integrate-and-fire~(IF) neuron model, without a refractory period, a membrane voltage of $v_t$ normalized to $[0, 1)$, a firing rate of $1$\,Hz, and the ODE discretized to a time-step of $\omega$\,s using zero-order hold~(ZOH).
The $\omega$ parameter essentially generalizes the spiking model to allow multiple spikes per time-step, and the IF restriction is lifted to allow arbitrary activation functions (including leaky neurons, and those with negative spikes such as $\tanh$).

\subsubsection{Scaling Properties}

We now state several important properties of this quantization algorithm (see supplementary for proofs).
For convenience, we assume the range of $f$ is scaled such that $|f(\cdot)| \le 1$ over the domain of valid inputs (this can also be achieved via clipping or saturation).

\paragraph{Zero-Mean Error} Supposing $v_{t-1} \sim \mathcal{U}[0, 1)$, the expected quantization error is $\mathds{E} \left[ \tilde{a}_t - a_t \right] = 0$.

\paragraph{Bounded Error} The total quantization error is bounded by $\left| \sum_{t \in T} \tilde{a}_t - a_t \right| < \omega^{-1}$ across \emph{any} consecutive slice of time-steps, $T$.
As a corollary, the signal-to-noise ratio~(SNR) of $\tilde{a}_t$ scales as $\Omega ( \omega )$, and this SNR may be further scaled by the time-constant of a lowpass filter (see section~\ref{sec:synapses}).

\paragraph{Bit-Width} The number of bits required to represent $\tilde{a}_t$ in binary is at most $\left\lceil \log_2 \left( \omega + 1 \right) \right\rceil$ if $f$ is non-negative (plus a sign bit if $f$ can be negative).

\paragraph{ANN Regime} As $\omega \rightarrow \infty$, $\tilde{a}_t \rightarrow a_t$, hence the activation function becomes equivalent to the ideal $f(\cdot)$.

\paragraph{SNN Regime} When $\omega \le 1$, the activation function becomes a conventional spiking neuron since it outputs either zero or a spike~($\omega^{-1}$) if $f$ is non-negative (and optionally a negative spike if $f$ is allowed to be negative).

\paragraph{Temporal Sparsity} 
The spike count scales as $\mathcal{O}(\omega)$.

To summarize, the choice of $\omega$ results in activities that require $\mathcal{O}(\log \omega)$ bits to represent, while achieving an SNR of $\Omega ( \omega )$ relative to the ideal function.
The effect of the algorithm is depicted in Figure~\ref{fig:dither} for various $\omega$.

\subsubsection{Backpropagation Training}

To train the network via backpropagation, we make the simplifying assumption that $(v_{t-1}, v_t) \sim \mathcal{U}[0, 1)$ are independent random variables, which implies that $\tilde{a}_t = a_t + \eta$ where $\eta \sim \mathcal{T}(-\omega^{-1}, \omega^{-1})$ is zero-mean noise with a symmetric triangular distribution (see supplementary).
This justifies assigning a gradient of zero to $\eta$.
The forward pass uses the quantized activation function to compute the true error for the current $\omega$, while the backward pass uses the gradient of $f$ (independently of $\omega$).
In summary, our scheme accounts for the temporal mechanisms of spike generation, but allows the gradient to skip over the sequence of operations that keep the total spike noise bounded by $\omega^{-1}$.

\subsection{Legendre Memory Unit}

As an example application of these methods we will use the Legendre Memory Unit~\citep[LMU;][]{voelker2019legendre}  -- a new type of Recurrent Neural Network~(RNN) that efficiently orthogonalizes the continuous-time history of some signal, $u(t) \in \mathbb{R}$, across a sliding window of length $\theta > 0$.
The network is characterized by the following $d$ coupled ODEs:
\begin{equation} \label{eq:delay-network}
    \theta \dot{\V{m}}(t) = \V{A}\V{m}(t) + \V{B}u(t)
\end{equation}
where $\V{m}(t) \in \mathbb{R}^d$ is a $d$-dimensional memory vector, and ($\V{A}$,~$\V{B}$) have a closed-form solution~\citep{voelker2019}:
\begin{equation} \label{eq:legendre-matrices}
\begin{aligned}
\V{A} = \left[ a \right]_{ij} \in \mathbb{R}^{d \times d} \text{,} \quad
a_{ij} &= \left(2i + 1\right)
\begin{dcases} 
  -1 & i < j \\
  (-1)^{i-j+1} & i \ge j
\end{dcases} \\
\V{B} = \left[ b \right]_i \in \mathbb{R}^{d \times 1} \text{,} \quad
b_i &= (2i + 1) (-1)^i \text{.} 
\end{aligned}
\end{equation}
The key property of this dynamical system is that $\V{m}$ represents sliding windows of $u$ via the \citet{legendre1782recherches} polynomials up to degree $d - 1$:
\begin{equation} \label{eq:legendre-polynomials}
\begin{aligned}
 u(t - \theta') &\approx \sum_{i=0}^{d-1} \mathcal{P}_{i} \left(\frac{\theta'}{\theta} \right) \, m_{i}(t) \text{,} \quad 0 \le \theta' \le \theta \\
 \mathcal{P}_i(r) &= (-1)^i \sum_{j=0}^i \begin{pmatrix}i \\ j\end{pmatrix} \begin{pmatrix}i + j \\ j\end{pmatrix} \left( -r \right)^j
\end{aligned}
\end{equation}
where $\mathcal{P}_i(r)$ is the $i^\text{th}$ shifted Legendre polynomial~\citep{rodrigues1816attraction}.
Thus, nonlinear functions of $\V{m}$ correspond to functions across windows of length $\theta$ projected onto the Legendre basis.

\paragraph{Discretization}

We map these equations onto the state of an RNN, $\V{m}_t \in \mathbb{R}^d$, given some input $u_t \in \mathbb{R}$, indexed at discrete moments in time, $t \in \mathbb{N}$:
\begin{equation} \label{eq:write-memory}
    \V{m}_t = f_m \left( \bar{\V{A}} \V{m}_{t-1} + \bar{\V{B}} u_t \right) 
\end{equation}
where ($\bar{\V{A}}$,~$\bar{\V{B}}$) are the ZOH-discretized matrices from equation~\ref{eq:legendre-matrices} for a time-step of $\bar{\theta}^{-1}$, such that $\bar{\theta}$ is the desired memory length expressed in discrete time-steps.
In the ideal case, $f_m(\cdot)$ should be the identity function.
For our hSNNs, we clip and then quantize $f_m(\cdot)$ using Algorithm~\ref{alg:quantizer}.


\paragraph{Architecture}

The LMU takes an input vector, $\V{x}_t$, and generates a hidden state.
The hidden state, $\V{h}_t \in \mathbb{R}^n$, and memory vector, $\V{m}_t \in \mathbb{R}^d$, correspond to the activities of two neural populations that we will refer to as the hidden neurons and memory neurons, respectively.
The hidden neurons mutually interact with the memory neurons in order to compute nonlinear functions across time, while dynamically writing to memory.
The state is a function of the input, previous state, and current memory:
\begin{equation}
    \V{h}_t = f_h \left( \V{W_x} \V{x}_t + \V{W_h} \V{h}_{t-1} + \V{W_m} \V{m}_t \right)
\end{equation}
where $f_h(\cdot)$ is some chosen nonlinearity---to be quantized using Algorithm~\ref{alg:quantizer}---and $\V{W_x}$, $\V{W_h}$, $\V{W_m}$ are learned weights.
The input to the memory is:
\begin{equation} \label{eq:memory-input}
    u_t = \transpose{\V{e_x}} \V{x}_t + \transpose{\V{e_h}} \V{h}_{t-1} + \transpose{\V{e_m}} \V{m}_{t - 1}
\end{equation}
where $\V{e_x}$, $\V{e_h}$, $\V{e_m}$ are learned vectors.

\subsection{Synaptic Filtering}
\label{sec:synapses}

SNNs commonly apply a synapse model to the weighted summation of spike-trains.
This filters the input to each neuron over time to reduce the amount of spike noise~\citep{dayan2001theoretical}.
The synapse is most commonly modelled as a lowpass filter, with some chosen time-constant $\tau$, whose effect is equivalent to replacing each spike with an exponentially decaying kernel, $h(t) = \tau^{-1} e^{-t / \tau}$ ($t \ge 0$).

By lowpass filtering the activities, the SNR of Algorithm~\ref{alg:quantizer} is effectively boosted by a factor of $\Omega(\tau)$ relative to the filtered ideal, since the filtered error becomes a weighted time-average of the quantization errors (see supplementary).
Thus, we lowpass filter the inputs into both $f_m(\cdot)$ and $f_h(\cdot)$.

To account for the temporal dynamics introduced by the application of a lowpass filter, \citet[][equation~4.7]{voelker2018improving} prove that the LMU's discretized state-space matrices, $(\bar{\V{A}}\text{,}\, \bar{\V{B}})$, should be exchanged with $(\bar{\V{A}}^H\text{,}\, \bar{\V{B}}^H)$:
\begin{equation} \label{eq:lmu-weights}
\begin{aligned}
    \bar{\V{A}}^H &= \frac{1}{1 - e^{-1 / \bar{\tau}}}\left(\bar{\V{A}} - e^{-1 / \bar{\tau}}I\right) \\
    \bar{\V{B}}^H &= \frac{1}{1 - e^{-1 / \bar{\tau}}} \bar{\V{B}}
\end{aligned}
\end{equation}
where $\bar{\tau}$ is the time-constant (in discrete time-steps) of the ZOH-discretized lowpass that is filtering the input to $f_m$.

To summarize, the architecture that we train includes a nonlinear layer ($\V{h}$) and a linear layer ($\V{m}$), each of which has synaptic filters.  The recurrent and input weights to $\V{m}$ are fixed to $\bar{\V{A}}^H$ and $\bar{\V{B}}^H$, and are not trained. All other connections are trained.

\subsection{SNR Scheduling}

To interpolate between ANN and SNN regimes, we set $\omega$ differently from one training epoch to the next, in an analogous manner to scheduling learning rates.
Since $\omega$ is exponential in bit-precision, we vary $\omega$ on a logarithmic scale across the interval $[\omega_h, \omega_l]$, where $\omega_h$ is set to achieve rapid convergence during the initial stages of training, and $\omega_l$ depends on the hardware and application.
Once $\omega = \omega_l$, training is continued until validation error stops improving, and only the model with the lowest validation loss during this fine-tuning phase is saved.

We found that this method of scheduling $\omega$ typically results in faster convergence rates versus the alternative of starting $\omega$ at its final value.
We also observe that the SNR of $f_m(\cdot)$ is far more critical than the SNR of $f_h(\cdot)$, and thus schedule the two differently (explained below).

\begin{table*}
\caption{Performance of RNNs on the sequential MNIST task.}
\label{tab:smnist}
\vskip 0.15in
\begin{center}
\begin{small}
\begin{sc}
\begin{tabular}{llllllllll}
\toprule
Network & Trainable & Weights & Nonlinearities & State & Levels & Steps & Test (\%) \\
\midrule
LSTM
 & 67850  
 & 67850
 & 384 sigmoid, 128 tanh
 & \textbf{256}
 & $2^{32}$
 & \textbf{784}
 & \textbf{98.5} \\
LMU
 & \textbf{34571}  
 & 51083  
 & \textbf{128 sigmoid}
 & \textbf{256}
 & $2^{32}$
 & \textbf{784}
 & 98.26 \\
hsLMU
 & \textbf{34571}  
 & 51083  
 & 128 LIF, 128 IF
 & 522
 & \textbf{2--5}
 & \textbf{784}
 & 97.26 \\
LSNN
 & 68210
 & \textbf{8185}
 & 120 LIF, 100 Adaptive
 & $\ge 550$
 & \textbf{2}
 & 1680  
 & 96.4 \\
\bottomrule
\end{tabular}
\end{sc}
\end{small}
\end{center}
\vskip -0.1in
\end{table*}

\section{Experiments}
\label{sec:experiments}

To facilitate comparison between the ``Long Short-Term Memory Spiking Neural Network''~(LSNN) from \citet{bellec2018long}, and both spiking and non-spiking LMUs~\citep{voelker2019legendre}, we consider the sequential MNIST~(sMNIST) task and its permuted variant~\citep[psMNIST;][]{le2015simple}.
For sMNIST, the pixels are supplied sequentially in a time-series of length $28^2$.
Thus, the network must maintain a memory of the relevant features while simultaneously computing across them in time.
For psMNIST, all of the sequences are also permuted by an unknown fixed permutation matrix, which distorts the temporal structure in the sequences and significantly increases the difficulty of the task.
In either case, the network outputs a classification at the end of each input sequence.
For the output classification, we take the argmax over a dense layer with 10 units, with incoming weights initialized using the Xavier uniform distribution~\citep{glorot2010understanding}.
The network is trained using the categorical crossentropy loss function (fused with softmax).


All of our LMU networks are built in Nengo~\citep{bekolay2014nengo} and trained using Nengo-DL~\citep{rasmussen2019nengodl}.
The 50k ``lost MNIST digits''~\citep{yadav2019cold}\footnote{This set does not overlap with MNIST's train or test sets.} are used as validation data to select the best model.
All sequences are normalized to $[-1, 1]$ via fixed linear transformation ($2x / 255 - 1$).
We use a minibatch size of $500$, and the Adam optimizer~\citep{kingma2014adam} with all of the default hyperparameters ($\alpha=0.001$, $\beta_1=0.9$, $\beta_2=0.999$).

To quantize the hidden activations, we use the leaky integrate-and-fire~(LIF) neuron model with a refractory period of 1 time-step and a leak of 10 time-steps (corresponding to Nengo's defaults given a time-step of 2\,ms), such that its response curve is normalized to $0 \le f_h(\cdot) < 1$.
The input to each LIF neuron is biased such that $f(x) = 0 \iff x \le 0$, and scaled such that $f(1) = e / (1 + e)$ (see supplementary).
During training, the $\omega$ for $f_h(\cdot)$ is interpolated across $[16, 1]$.
Thus, the hidden neurons in the fully trained networks are conventional (1-bit) spiking neurons.

To quantize the memory activations, we use $f_m(x) = x\text{.clip}(-1, +1)$, which is analogous to using IF neurons that can generate both positive and negative spikes.
To maintain accuracy, the $\omega$ for $f_m(\cdot)$ is interpolated across $[32, 
2]$ for sMNIST, and across $[4080, 
255]$ 
for psMNIST.
We provide details regarding the effect of these choices on the number of possible activity levels for the memory neurons, and discuss the impact this has on MAC operations as well as the consequences for energy-efficient neural networks.

The synaptic lowpass filters have a time-constant of 200 time-steps for the activities projecting into $f_m(\cdot)$, and 10 time-steps for the activities projecting into $f_h(\cdot)$.
The output layer also uses a 10 time-step lowpass filter.
We did not experiment with any other choice of time-constants.

All weights are initialized to zero, except: $\V{e_x}$ is initialized to $1$, $\V{W}_\V{m}$ is initialized using the Xavier normal distribution~\citep{glorot2010understanding},
and $(\bar{\V{A}}^H\text{,}\, \bar{\V{B}}^H)$ are initialized according to equation~\ref{eq:lmu-weights} and left untrained.
L2-regularization ($\lambda = 0.01$) is added to the output vector.
We did not experiment with batch normalization, layer normalization, dropout, or any other regularization techniques.

\begin{table*}
\caption{Performance of RNNs on the permuted sequential MNIST task.}
\label{tab:psmnist}
\vskip 0.15in
\begin{center}
\begin{small}
\begin{sc}
\begin{tabular}{lllllll}
\toprule
Network & Trainable & Weights & Nonlinearities & Bit-Width & Significant Bits & Test (\%) \\
\midrule
LSTM
 & 163610  
 & \textbf{163610}
 & 600 sigmoid, 200 tanh
 & 32
 & N/A
 & 89.86 \\
LMU
 & \textbf{102027}  
 & 167819  
 & \textbf{212 tanh}
 & 32
 & N/A
 & \textbf{97.15} \\
hsLMU
 & 102239  
 & 168031  
 & 212 LIF, 256 IF
 & \textbf{3.74}  
 & \textbf{1.26}
 & 96.83 \\
\bottomrule
\end{tabular}
\end{sc}
\end{small}
\end{center}
\vskip -0.1in
\end{table*}

\subsection{Sequential MNIST}

\subsubsection{State-of-the-Art}

The LSTM and LSNN results shown in Table~\ref{tab:smnist} have been extended from \citet[][Tables~S1 and~S2]{bellec2018long}.
We note that these two results (98.5\% and 96.4\%) represent the best test accuracy among 12 separately trained models, without any validation.  
The mean test performance across the same 12 runs are 79.8\% and 93.8\% for the LSTM and LSNN, respectively.

The LSTM consists of only 128 ``units,'' but is computationally and energetically intensive since it maintains a 256-dimensional vector of 32-bit activities that are multiplied with over 67k weights.
The LSNN improves this in two important ways.
First, the activities of its 220 neurons are all one bit (i.e.,~spikes).
Second, the number of parameters are pruned down to just over 8k weights.
Thus, each time-step consists of at most 8k synaptic operations that simply add a weight to the synaptic state of each neuron, followed by local updates to each synapse and neuron model.

However, the LSNN suffers from half the throughput (each input pixel is presented for two time-steps rather than one), a latency of 112 additional time-steps to accumulate the classification after the image has been presented (resulting in a total of $2 \times 28^2 + 112 = 1680$ steps), and an absolute 2.1\% decrease in test accuracy.
In addition, at least 550 state-variables (220 membrane voltages, 100 adaptive thresholds, 220 lowpass filter states, 10 output filters states, plus state for an optional delay buffer attached to each synapse) are required to maintain memory between time-steps.
The authors state that the input to the LSNN is preprocessed using 80 more neurons that fire whenever the pixel crosses over a fixed value associated with each neuron, to obtain ``somewhat better performance.''

\begin{figure}[ht!]
\begin{center}
\centerline{\includegraphics[width=\columnwidth]{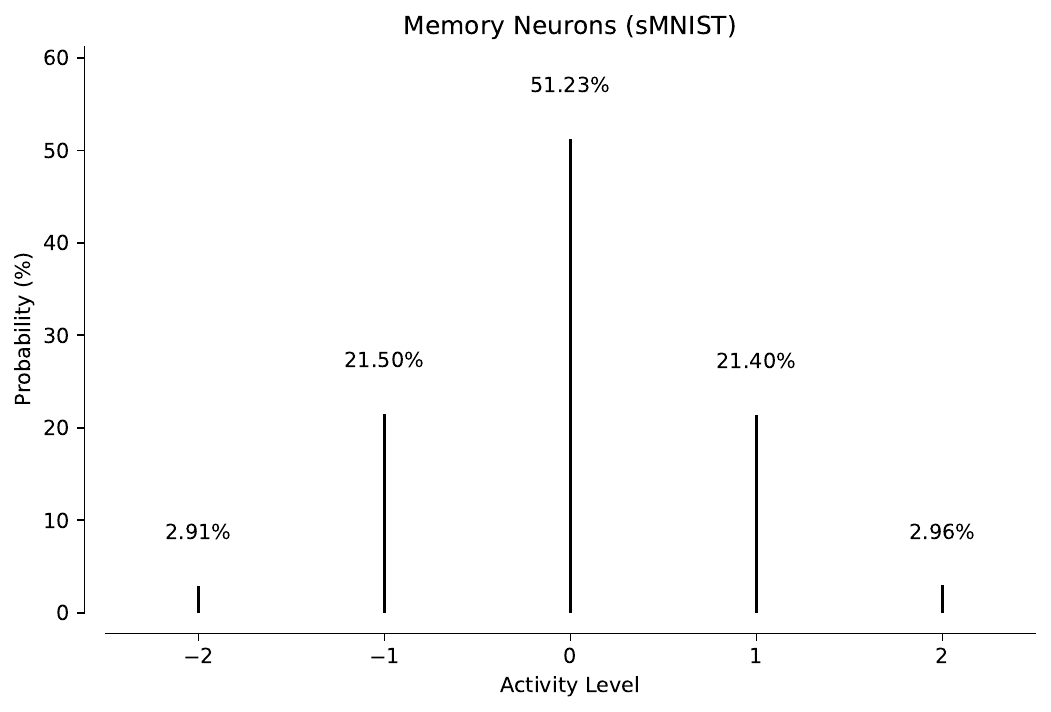}}
\centerline{\includegraphics[width=\columnwidth]{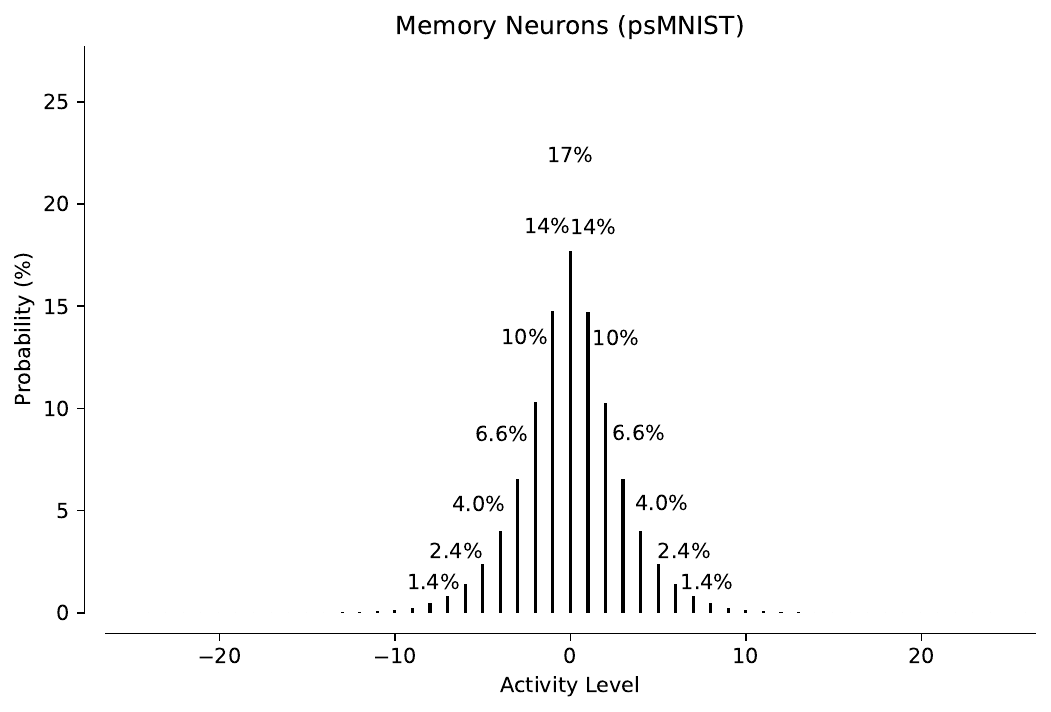}}
\caption{
Distribution of activity levels for the memory neurons, $f_m(\cdot)$, in the hsLMU network solving the sMNIST task (Top; see Table~\ref{tab:smnist}) and the psMNIST task (Bottom; see Table~\ref{tab:psmnist}).
}
\label{fig:smnist_lmu_spiking_memory}
\end{center}
\vspace{-2.5em}
\end{figure}

\subsubsection{Non-Spiking LMU}

The non-spiking LMU is the Nengo implementation from \citet{voelker2019legendre}\footnote{\url{https://www.nengo.ai/nengo-dl/examples/lmu.html}} with $n=128$ and $d=128$, the sigmoid activation chosen for $f_h(\cdot)$, and a trainable bias vector added to the hidden neurons.

This network obtains a test accuracy of 98.26\%, while using only 128 nonlinearities, and training nearly half as many weights as the LSTM or LSNN.
However, the MAC operations are still a bottleneck, since each time-step requires multiplying a 256-dimensional vector of 32-bit activities with approximately 51k weights (including $\bar{\V{A}}^H$ and $\bar{\V{B}}^H$).

\subsubsection{Hybrid-Spiking LMU}
\label{sec:smnist-lmu}

To simplify the MAC operations, we quantize the activity functions and filter their inputs (see section~\ref{sec:methods}).
We refer to this as a ``hybrid-spiking LMU''~(hsLMU) since the hidden neurons are conventional (i.e., one-bit) spiking LIF neurons, but the memory neurons can assume a multitude of distinct activation levels by generating multiple spikes per time-step.

By training until $\omega = 2$ for $f_m(\cdot)$, each memory neuron can spike at 5 different activity levels (see Figure~\ref{fig:smnist_lmu_spiking_memory};~Top).
We remark that the distribution is symmetric about zero, and ``prefers'' the zero state~(51.23\%), since equation~\ref{eq:delay-network} has exactly one stable point: $\V{m}(t) = \V{0}$.
As well, the hidden neurons spike only 36.05\% of the time.
As a result, the majority of weights are not needed on any given time-step.
Furthermore, when a weight is accessed, it is simply added for the hidden activities, or multiplied by an element of $\{-2, -1, +1, +2\}$ for the memory activities.

These performance benefits come at the cost of a 1\% decrease in test accuracy, and additional state and computation---local to each neuron---to implement the lowpass filters and Algorithm~\ref{alg:quantizer}.
Specifically, this hsLMU requires 522 state-variables (256 membrane voltages, 256 lowpass filters, and 10 output filters).
This network outperforms the LSNN, does not sacrifice throughput nor latency, and does not require special preprocessing of the input data.

\subsection{Permuted Sequential MNIST}

\subsubsection{State-of-the-Art}

Several RNN models have been used to solve the psMNIST task~\citep{chandar2019towards}, with the highest accuracy of 97.15\% being achieved by an LMU~\citep{voelker2019legendre} of size $n = 212$, $d = 256$.
The LMU result, and the LSTM result from \citet{chandar2019towards}, are reproduced in Table~\ref{tab:psmnist}.

\begin{figure*}
\begin{center}
\centerline{\includegraphics[width=\textwidth]{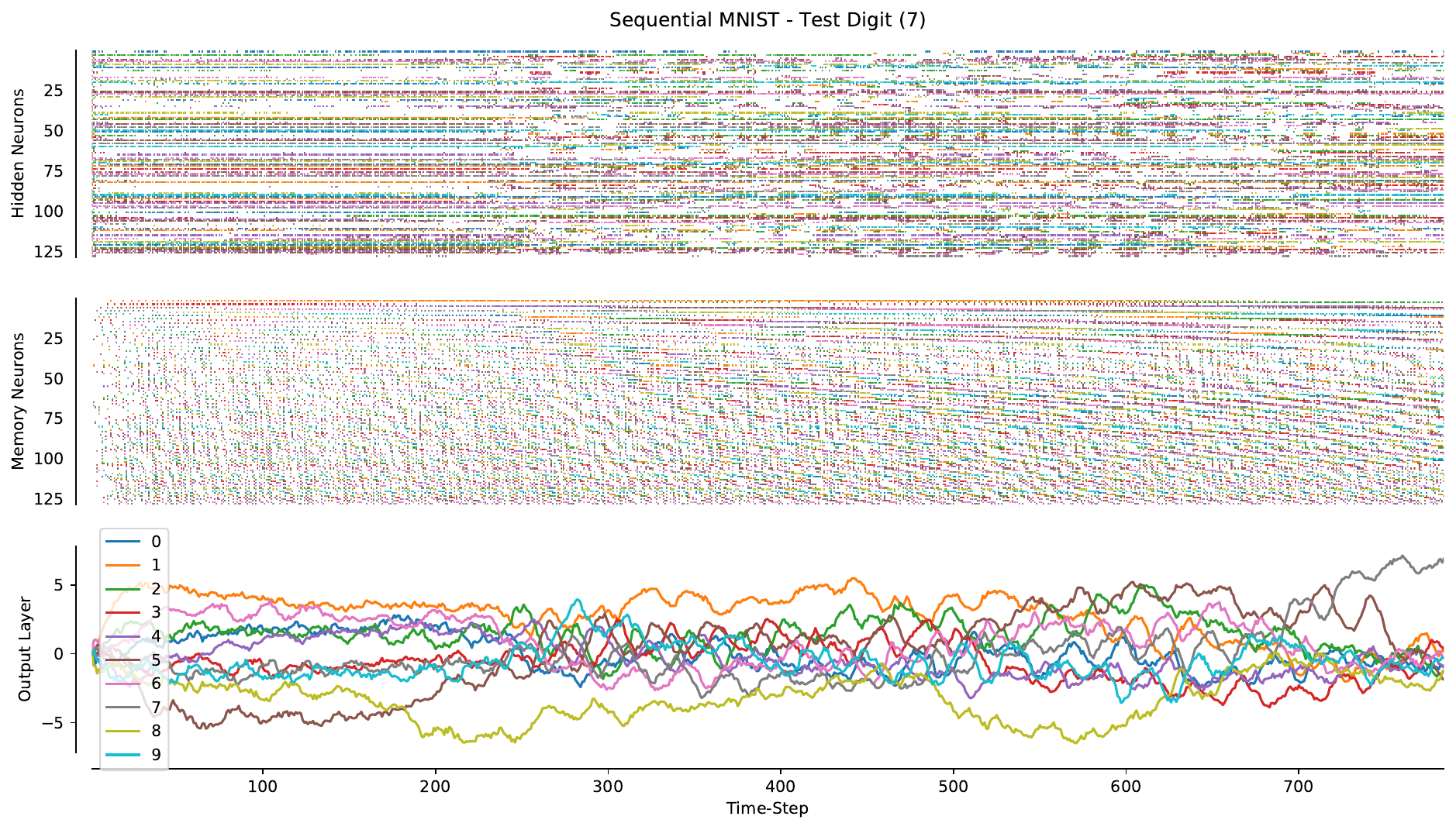}}
\vskip 0.1in
\centerline{\includegraphics[width=\textwidth]{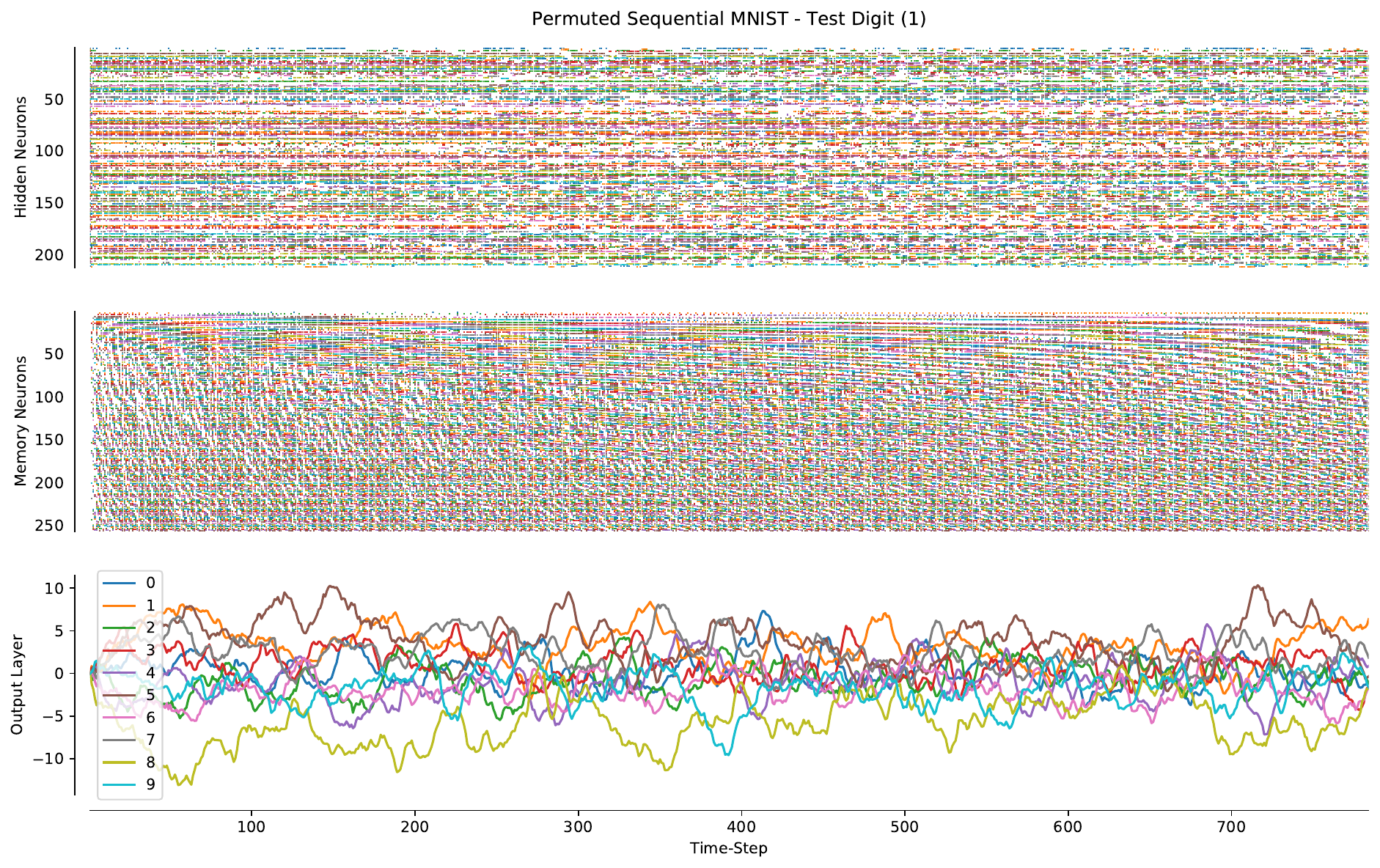}}
\caption{
An example of each hybrid-spiking Legendre Memory Unit~(hsLMU) network producing the correct classification given a test digit for the sMNIST task (Top; see Table~\ref{tab:smnist}) and the psMNIST task (Bottom; see Table~\ref{tab:psmnist}).
The recurrent network consists of one-bit spiking LIF neurons (representing $\V{h}_t$) coupled with multi-bit spiking IF neurons (representing $\V{m}_t$).
Classifications are obtained by taking an argmax of the output layer on the final time-step of each sequence.
}
\label{fig:smnist_lmu_spiking_traces}
\end{center}
\end{figure*}

\subsubsection{Hybrid-Spiking LMU}

We consider the same network from section~\ref{sec:smnist-lmu}, scaled up to $n = 212$ and $d = 256$.
Consistent with the previous hsLMU, the hidden neurons are spiking LIF, and the memory neurons are multi-bit IF neurons that can generate multiple positive or negative spikes per step.
In particular, by training until $\omega = 255$ for $f_m(\cdot)$, each memory neuron can spike between -24 and +26 times (inclusive) per step for a total of 50 distinct activity levels, which requires 6 bits to represent.

Again, the distribution of memory activities are symmetric about zero, and 17.71\% of the time the neurons are silent.
The 1-bit hidden neurons spike 40.24\% of the time.
We note that the hsLMU uses 212 more parameters than the LMU from \citet{voelker2019legendre}, as the latter does not include a bias on the hidden nonlinearities.

To quantify the performance benefits of low-precision activities, we propose the following two metrics.
The first is the worst-case number of bits required to communicate the activity of each neuron, in this case $1$ for the hidden neurons and $6$ for the memory neurons, which has a weighted average of approximately $3.74$ bits.
The second is the number of bits that are significant (i.e., after removing all of the trailing zero bits, and including a sign bit for negative activities), which has a weighted average of approximately $1.26$ bits.

The ``bit-width'' metric is useful for determining the worst-case volume of spike traffic on hardware where the size of the activity vectors are user-configurable~\citep{furber2014spinnaker, liu2018memory}, and for hardware where the quantization of activities leads to quadratic improvements in silicon area and energy requirements~\citep{mckinstry2018discovering}.
The ``significant bits'' metric reflects how many significant bits are multiplied with each weight, which is important for hardware where bit-flips in the datapath correlate with energy costs~\citep{liimproving2019}, or hardware that is optimized for integer operands close to zero.
For instance, a value of 1 for this metric would imply that each MAC, on average, only needs to accumulate its weight (i.e., no multiply is required).
These performance benefits come at the cost of a 0.32\% decrease in test accuracy, which still outperforms all other RNNs considered by \citet{chandar2019towards, voelker2019legendre} apart from the LMU, while using comparable resources and parameter counts.

Interestingly, for the sMNIST network in section~\ref{sec:smnist-lmu}, the bit-width metric is exactly 2 (as there are an equal number of hidden (1-bit) and memory (3-bit) neurons). The significant bits for that network is 0.58, because a majority of the neurons are inactive on each time step. 

\section{Discussion}

Although the biological plausibility of a neuron that can output more than one spike ``at once'' is questionable, it is in fact mathematically equivalent to simulating the neuron with a time-step of $\omega$ and bundling the spikes together~\citep{krithivasan2019dynamic}. Consequently, all of the networks we consider here can be implemented in 1-bit spiking networks, although with an added time cost.
This is similar to the LSNN's approach of simulating the network for two time-steps per image pixel, but does not incur the same cost in throughput.
Alternatively, a space cost can be paid by replicating the neuron $\omega$ times and uniformly spacing the initial $v_0$ (not shown).
Likewise, negative spikes are a more compact and efficient alternative to duplicating the neurons and mirroring their activation functions.

Our methods are convenient to apply to the LMU because equation~\ref{eq:lmu-weights} accounts for the dynamics of the lowpass filter, and the $\V{m}_t$ vector naturally prefers the zero (i.e., silent) state.
At the same time, it is a challenging test for the theory, since we do not train the LMU matrices, which are primarily responsible for accuracy on psMNIST~\citep{voelker2019legendre}, and RNNs tend to accumulate and propagate their errors over time.
Notably, the method of Algorithm~\ref{alg:quantizer} can be applied to other neural network architectures, including feed-forward networks.

\section{Conclusions}

We have presented a new algorithm and accompanying methods that allow interpolation between spiking and non-spiking networks.  This allows the training of hSNNs, which can have mixtures of activity quantization, leading to computationally efficient neural network implementations.  We have also shown how to incorporate standard SNN assumptions, such as the presence of a synaptic filter. 

We demonstrated the technique on the recently proposed LMU, and achieved better than state-of-the-art results on sMNIST than a spiking network.
Additionally, on the more challenging psMNIST task the reported accuracy of the spiking network is better than any non-spiking RNN apart from the original LMU~\citep{chandar2019towards, voelker2019legendre}.

However, our focus here is not on accuracy per se, but efficient computation. In this context, the training procedure enables us to leverage the accuracy of ANNs and the energy efficiency of SNNs by scheduling training to evaluate a series of networks in between these two extremes.  In the cases we considered, we reduced the activity to 2--6 bits on average, saving at least 26 bits over the standard LMU implementation with minimal impact on accuracy.  While difficult to convert these metrics to energy savings in a hardware-agnostic manner, such optimizations can benefit both spiking and non-spiking architectures.

We anticipate that techniques like those we have outlined here will become more widely used as the demands of edge computing continue to grow.  In such power-constrained contexts, extracting as much efficiency as possible, while retaining sufficient accuracy, is central to the efforts involved in co-designing both algorithms and hardware for neural network workloads.


\newpage

\bibliography{main}
\bibliographystyle{icml2020}

\newpage
\onecolumn

\title{A Spike in Performance -- Supplementary}
\author{Aaron R. Voelker, Daniel Rasmussen, and Chris Eliasmith}
\date{February 10, 2020}

\maketitle
\thispagestyle{empty}  

\setcounter{section}{0}
\setcounter{algorithm}{0}

\section{Introduction}

For clarity, we restate Algorithm~1 from the main article with some additional annotations.
This algorithm quantizes an activation function, $f$, given a parameter $\omega > 0$ that acts as a time-step.
For convenience, we assume the range of $f$ is scaled such that $|f(\cdot)| \le 1$ over the domain of valid inputs.
This may be accomplished by fusing a normalization term with the weights, or by clipping or saturating the nonlinearity.
The ideal output of the activation function at step $t$ is defined as $a_t = f(x_t)$.
The algorithm maintains one scalar state-variable per neuron, $v_{t-1} \mapsto v_{t}$ (i.e.,~from one step to the next), initialized by sampling $v_0 \sim \mathcal{U}[0, 1)$.

\centerline{
\begin{minipage}{.42\linewidth}
\begin{algorithm}[H]
   \caption{\newline Temporally-Diffused Quantizer ($f$; $\omega$)}
\begin{algorithmic}
   \STATE {\bfseries Input:} $x_t$
   \STATE {\bfseries State:} $v_t$
   \STATE {\bfseries Output:} $\tilde{a}_t$
   \STATE $s_t \leftarrow v_{t-1} + \overbrace{f(x_t) \times \omega}^{c_t}$
   \STATE $k_t \leftarrow \left \lfloor s_t \right \rfloor$
   \STATE $v_t \leftarrow s_t - k_t$
   \STATE $\tilde{a}_t \leftarrow k_t / \omega$
\end{algorithmic}
\end{algorithm}
\vspace{0.5em}
\end{minipage}
}
\noindent
Let $c_t = f(x_t) \times \omega$, since this term will appear frequently, and (as we will see) corresponds to the expected spike count.
We'll also regularly use the facts that $k_t = \floor{s_t} = \floor{v_{t-1} + f(x_t) \times \omega} = \floor{v_{t-1} + c_t}$, $f(x_t) = a_t = c_t / \omega$, and $0 \le v_t < 1$ for all $t \in \mathbb{N}$.

Section~\ref{sec:proofs} restates and then proves each property from the main article (and in a slightly different order).
Section~\ref{sec:uniform-voltage} provides some empirical justification for the useful assumption that the neurons' voltages, $v_t$, remain uniformly distributed in practice.
Section~\ref{sec:lif} details how we scale and bias the leaky integrate-and-fire~(LIF) response curve.

\newpage

\section{Statements and Proofs}
\label{sec:proofs}


\begin{theorem}[Zero-Mean Error] \label{thm:error}
Supposing $v_{t-1} \sim \mathcal{U}[0, 1)$, the expected quantization error is $\mathds{E} \left[ \tilde{a}_t - a_t \right] = 0$.
\end{theorem}

\begin{proof}
Let $b_t = 1 - (c_t - \floor{c_t}) = \floor{1 + c_t} - c_t$ correspond to the point at which $v_{t-1} + c_t$ floors down to $\floor{c_t}$ for $v_{t-1} < b_t$, and otherwise $\floor{1+c_t}$ for $v_{t-1} \ge b_t$.
By the law of the unconscious statistician:

\begin{align*}
\mathds{E} \left[ \tilde{a}_t - a_t \right] &= \int_{0}^1 \left( \tilde{a}_t - a_t \right) dv_{t - 1} \\
&= \omega^{-1} \int_{0}^1 \left(\floor{v_{t-1} + c_t} - c_t\right) dv_{t-1} \\
&= \omega^{-1} \left[ \int_{0}^{b_t} \left(\floor{v_{t-1} + c_t} - c_t\right) dv_{t-1} + \int_{b_t}^{1} \left(\floor{v_{t-1} + c_t} - c_t\right) dv_{t-1} \right] \\
&= \omega^{-1} \left[ \int_{0}^{b_t} \left(\floor{c_t} - c_t\right) dv_{t-1} + \int_{b_t}^{1} \left(\floor{1 + c_t} - c_t\right) dv_{t-1} \right] \\
&= \omega^{-1} \left[ b_t (b_t - 1) + (1 - b_t) b_t \right] \\
&= 0 \text{.}
\end{align*}
\end{proof}

\begin{corollary}[Expected Spike Count]
$\mathds{E} \left[ k_t \right] = c_t$,
hence $c_t$ corresponds to the expected number of spikes (positive or negative) across a time window of length $\omega$.
\end{corollary}

\begin{proof}
By linearity of expectation, and since $\mathds{E} \left[ \tilde{a}_t \right] = a_t = f(x_t)$,
\begin{equation*}
\mathds{E} \left[ k_t \right] = \mathds{E} \left[ k_t / \omega \right] \times \omega = a_t \times \omega = c_t \text{.}
\end{equation*}
\end{proof}

\begin{corollary}[Temporal Sparsity]
The spike count scales as $\mathcal{O}(\omega)$.
\end{corollary}

\begin{proof}
Since $|f(\cdot)| \le 1$ is $\mathcal{O}(1)$, we have $k_t$ scaling as $\mathcal{O}(c_t) = \mathcal{O}(\omega)$.
\end{proof}

\begin{theorem}[Bounded Error] \label{thm:bounds}
The total quantization error is bounded by $\left| \sum_{t \in T} \tilde{a}_t - a_t \right| < \omega^{-1}$ across \emph{any} consecutive slice of time-steps, $T$.
\end{theorem}

\begin{proof}
Suppose $T$ is the discrete interval $[i + 1, j]$, for natural numbers $i < j$. Using the fact that $\tilde{a}_t = k_t / \omega = (s_t - v_t) / \omega = (v_{t-1} + c_t - v_t) / \omega$, we obtain a telescoping sum:
\begin{align*}
    \sum_{t \in T} \tilde{a}_t - a_t &= \omega^{-1} \left( \sum_{t=i+1}^{j} (v_{t-1} + c_t - v_t) - c_t \right) \\
    &= \omega^{-1} \left( \sum_{t=i+1}^{j} v_{t-1} - v_t \right) \\
    &= \omega^{-1} (v_i - v_j) \text{.}
\end{align*}
Since $0 \le (v_i, v_j) < 1$, this implies $|v_i - v_j| < 1$ and therefore $\left| \sum_{t \in T} \tilde{a}_t - a_t \right| < \omega^{-1}$.
\end{proof}

\begin{corollary}[ANN Regime]
As $\omega \rightarrow \infty$, $\tilde{a}_t \rightarrow a_t$, hence the activation function becomes equivalent to the ideal $f(\cdot)$.
\end{corollary}

\begin{proof}
As a corollary to Theorem~\ref{thm:bounds}  with $T = \{ t\}$, $\tilde{a}_t \rightarrow a_t$ in the limit of $\omega \rightarrow \infty$.
\end{proof}

\begin{corollary}[Backpropagation Training] \label{cor:gradient}
Assuming $(v_{t-1}, v_t) \sim \mathcal{U}[0, 1)$ are independent random variables, 
we have $\tilde{a}_t = a_t + \eta$ where $\eta \sim \mathcal{T}(-\omega^{-1}, \omega^{-1})$ is zero-mean noise with a symmetric triangular distribution.
This justifies assigning a gradient of zero to $\eta$ during the backwards pass of the backpropagation algorithm.
\end{corollary}

\begin{proof}
Consider any $t$, and apply Theorem~\ref{thm:bounds} with $T = \{ t\}$ to get $\eta = \tilde{a}_t - a_t = \omega^{-1} \left(v_{t - 1} - v_t\right)$, which is the scaled difference between two independently distributed uniform random variables.
This results in the standard triangular distribution, symmetric about zero, and divided by $\omega$.
We briefly note that the covariance in $\eta$ from one time-step to the next is: $$\text{cov}( \omega^{-1} \left(v_{t - 1} - v_t\right), \omega^{-1} \left(v_{t} - v_{t+1}\right)) = \omega^{-2} \left(\frac{1}{4} - \frac{1}{4} - \frac{1}{3} + \frac{1}{4}\right) = -1/\left(12\omega^2\right) \text{.}$$
\end{proof}

\begin{corollary}[Signal-to-Noise Ratio] \label{cor:snr}
Supposing $\mathds{E} \left[ \tilde{a}_t - a_t \right] = 0$, the signal-to-noise ratio~(SNR) of $\tilde{a}_t$ scales as $\Omega ( \omega )$.
\end{corollary}

\begin{proof}
Since the signal has zero-mean error (by assumption), the SNR, defined as the ratio of the signal to the standard deviation of its error, 
is:
$$\text{SNR} = \frac{a_t}{\sqrt{ \mathds{E} \left[ \left( \tilde{a}_t - a_t \right)^2 \right]}} \text{.}$$
Popoviciu's inequality, applied to Theorem~\ref{thm:bounds} with $T = \left\{ t \right\}$, gives the bound: $$\sqrt{ \mathds{E} \left[ \left( \tilde{a}_t - a_t \right)^2 \right]} \le \frac{1}{2} \left( \omega^{-1} - (-\omega^{-1}) \right) = \omega^{-1} \text{.}$$
Hence, $\text{SNR} \ge a_t \omega = \Omega(\omega)$. 
Furthermore, we remark that assuming $v_{t-1} \sim \mathcal{U}[0, 1)$ leads to a stronger bound (by a factor of $2$).
In particular, using this assumption, the variance term can be derived similarly to the zero-mean error, while using the fact that $0 < b_t \le 1$ to establish a tighter upper-bound:
\begin{align*}
\mathds{E} \left[ \left(\tilde{a}_t - a_t\right)^2 \right] &= \int_{0}^1 \left( \tilde{a}_t - a_t \right)^2 dv_{t - 1} \\
&= \omega^{-2} \left[ \int_{0}^{b_t} \left(\floor{c_t} - c_t\right)^2 dv_{t-1} + \int_{b_t}^{1} \left(\floor{1 + c_t} - c_t\right)^2 dv_{t-1} \right] \\
&= \omega^{-2} \left[ b_t (b_t - 1)^2 + (1 - b_t) b_t^2 \right] \\
&= \omega^{-2} \left( b_t (1 - b_t) \right) \\
&\le \left(\frac{1}{2\omega}\right)^2 \text{.}
\end{align*}
Hence,
$\text{SNR} \ge 2 a_t \omega = \Omega(\omega) \text{.}$
\end{proof}


\begin{theorem}[Synaptic Filtering] \label{thm:synapse}
The SNR from Corollary~\ref{cor:snr} may be further scaled by the time-constant of a lowpass filter.
Specifically, supposing $\mathds{E} \left[ \tilde{a}_t - a_t \right] = 0$, the SNR is $\ge \omega / (1 - e^{-1 / \bar{\tau}}) = \Omega(\omega \bar{\tau})$ where $\bar{\tau} = \tau / dt$ is the discrete time-constant of the zero-order hold~(ZOH) lowpass filter applied to $\tilde{a}$.
\end{theorem}

\begin{proof}
Intuitively, when $\tilde{a}$ is filtered by a first-order lowpass filter, the quantization errors cancel out in a similar manner to Theorem~\ref{thm:bounds}, which boosts the SNR by the time-constant of the filter.
While it is possible to give a much more general result, we instead prove this in a direct and self-contained manner.
We note that by linearity of convolution, it does not functionally matter whether we apply the filter before or after the connection weight-matrix.

First, we simply state that ZOH-discretizing a first-order lowpass filter with a continuous-time impulse response of $h(t) = \tau^{-1} e^{-t/\tau}$ ($t \ge 0$), with a time-step of $dt$, results in the discrete-time impulse response of: $$h_t = \left(1 - e^{-1/\bar{\tau}}\right) e^{-t/\bar{\tau}}  \text{,} \quad t \in \mathbb{N} \text{.}$$
For completeness, we note that, in the Laplace domain, $H(s) = 1 / (\tau s + 1)$ and: $$H(z) = \frac{1 - e^{-1/\bar{\tau}}}{z - e^{-1/\bar{\tau}}} \text{.}$$
Second, the quantity of interest---the filtered error at step $t$---is the following discrete-time convolution:
$$\left( \left( \tilde{a} - a \right) \ast h\right)_t \overset{\text{DEF}}{=} \sum_{i=0}^\infty \left( \tilde{a}_{t-i} - a_{t-i} \right) h_i \text{,}$$
where negative indices are handled appropriately by setting those values to zero.
Third, we note that $h_{i-1} > h_i$ for all $i > 0$.
Fourth, $1 - e^{-y} < y$ for any $y > 0$ ($y = 1 / \bar{\tau}$ in particular).
We are now ready to bound the filtered error:
\begin{align*}
\left| \left( \left( \tilde{a} - a \right) \ast h \right)_t \right| &= \left| \sum_{i=0}^\infty \left( \tilde{a}_{t-i} - a_{t-i} \right) h_i \right| \\
&= \omega^{-1} \left| \sum_{i=0}^\infty \left( v_{t-i-1} - v_{t-i} \right) h_i \right| \\
&= \omega^{-1} \left| -h_0 v_t + \sum_{i=1}^\infty \left(h_{i-1} - h_i\right) v_{t-i} \right| \\
&< \omega^{-1} \max\left( h_0, \sum_{i=1}^\infty h_{i-1} - h_i \right) \\
&= \omega^{-1} \max\left( \left(1-e^{-1/\bar{\tau}}\right), \left(1 - e^{-1/\bar{\tau}}\right) \sum_{i=1}^\infty e^{-(i-1)/\bar{\tau}} - e^{-i/\bar{\tau}} \right) \\
&= \omega^{-1} \left(1-e^{-1/\bar{\tau}}\right) \max\left( 1, \left(e^{1/\bar{\tau}} - 1\right) \sum_{i=1}^\infty e^{-i/\bar{\tau}} \right) \\
&= \omega^{-1} \left(1-e^{-1/\bar{\tau}}\right) \max\left( 1, \left(e^{1/\bar{\tau}} - 1\right) \frac{1}{e^{1/\bar{\tau}} - 1} \right) \\
&= \omega^{-1} \left(1-e^{-1/\bar{\tau}}\right) \\
&< \frac{1}{\omega \bar{\tau}} \text{.}
\end{align*}
We remark that the third line is equivalent in the $\mathcal{Z}$-domain to multiplying $(z^{-1} - 1)H(z)$ with the $\mathcal{Z}$-transform of $v$.
Although it is once again possible to obtain tighter bounds by assuming uniform voltages,
simply applying Popoviciu's inequality produces the filtered $\text{SNR} \ge (a \ast h)_t \, \omega / \left(1-e^{-1/\bar{\tau}}\right) = \Omega(\omega \bar{\tau})$.
\end{proof}

\begin{lemma}[Bit-Width] \label{lemma:bit-width}
The number of bits required to represent $\tilde{a}_t$ in binary is at most $\left\lceil \log_2 \left( \omega + 1 \right) \right\rceil$ if $f$ is non-negative (plus a sign bit if $f$ can be negative).
\end{lemma}

\begin{proof}
If $f$ is non-negative, then $k_t$ can take on $\omega + 1$ different values (specifically, $k_t \in [0, \omega]$).
Representing $\omega + 1$ distinct values in binary requires $\left\lceil \log_2 \left( \omega + 1 \right) \right\rceil$ bits in the worst case.
If $f$ can be negative, then $k_t$ can take on discrete values across the interval $[-\omega, \omega]$ and thus an additional sign bit is required.
\end{proof}

\begin{corollary}[SNN Regime]
When $\omega \le 1$, the activation function becomes a conventional spiking neuron since it outputs either zero or a spike~($\omega^{-1}$) if $f$ is non-negative (and optionally a negative spike if $f$ is allowed to be negative).
\end{corollary}

\begin{proof}
As a corollary to Lemma~\ref{lemma:bit-width}, a choice of $\omega \le 1$ results in activities that can be represented using at most one bit if $f$ is non-negative (and optionally a sign bit if $f$ is allowed to be negative).
Hence the neuron model can only signal events (i.e.,~``spikes'') for non-negative response functions, and optionally negative spikes for signed response functions.
\end{proof}

\begin{minipage}{\linewidth}
\begin{figure}[H]
\includegraphics[width=\columnwidth]{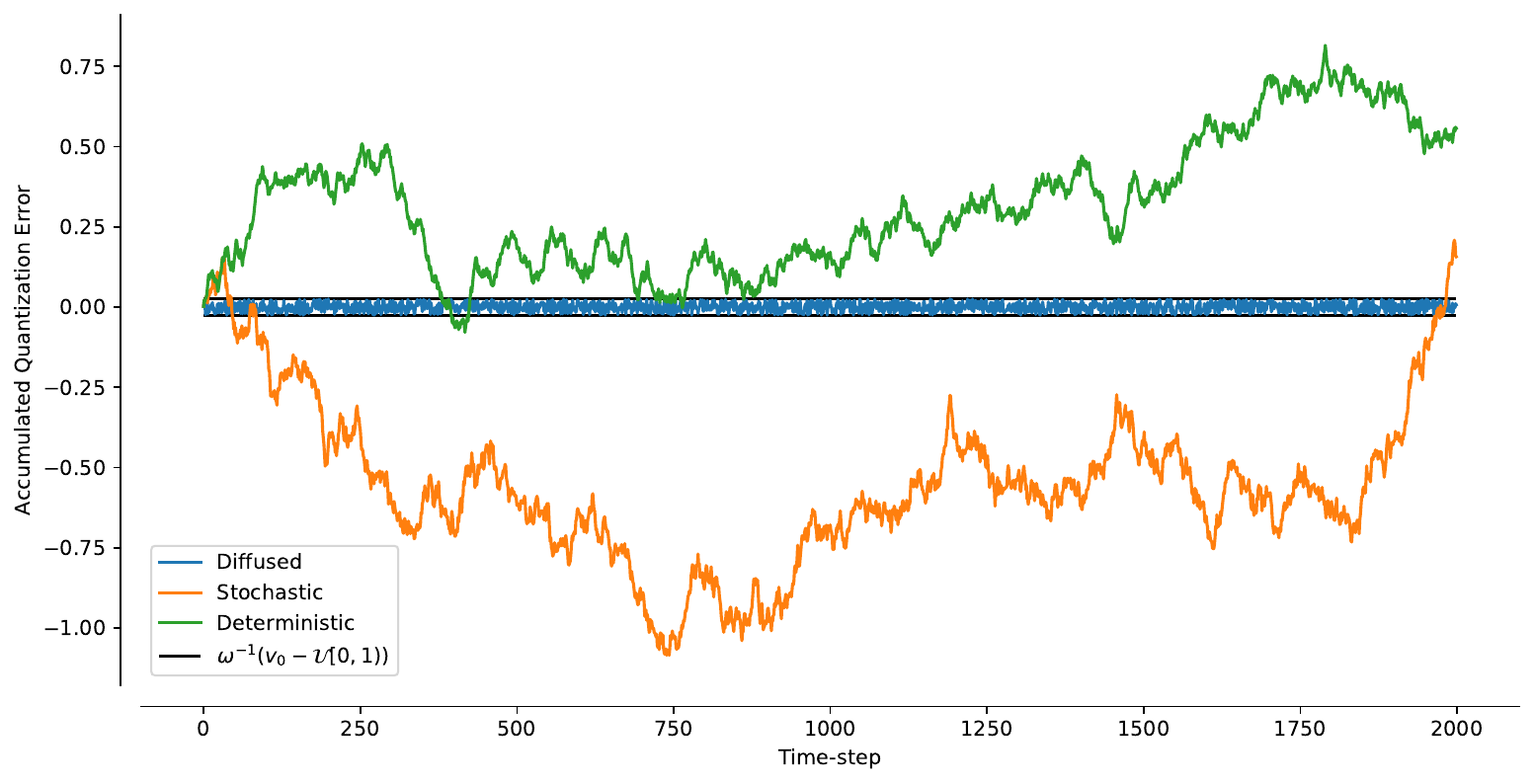}
\caption{ \label{fig:bounds}
Numerical validation of Theorem~\ref{thm:bounds} given a random input signal, and compared to the accumulated error for two alternative quantizers: a deterministic quantizer that does not temporally-diffuse its error, and a non-deterministic quantizer that uses stochastic rounding.
The voltage is initialized to $v_0 = 0.5$ in this case.
It is especially important to keep the total quantization error bounded when considering Recurrent Neural Networks~(RNNs), since they tend to accumulate and propagate their errors over time.
}
\end{figure}

\begin{figure}[H]
\includegraphics[width=\columnwidth]{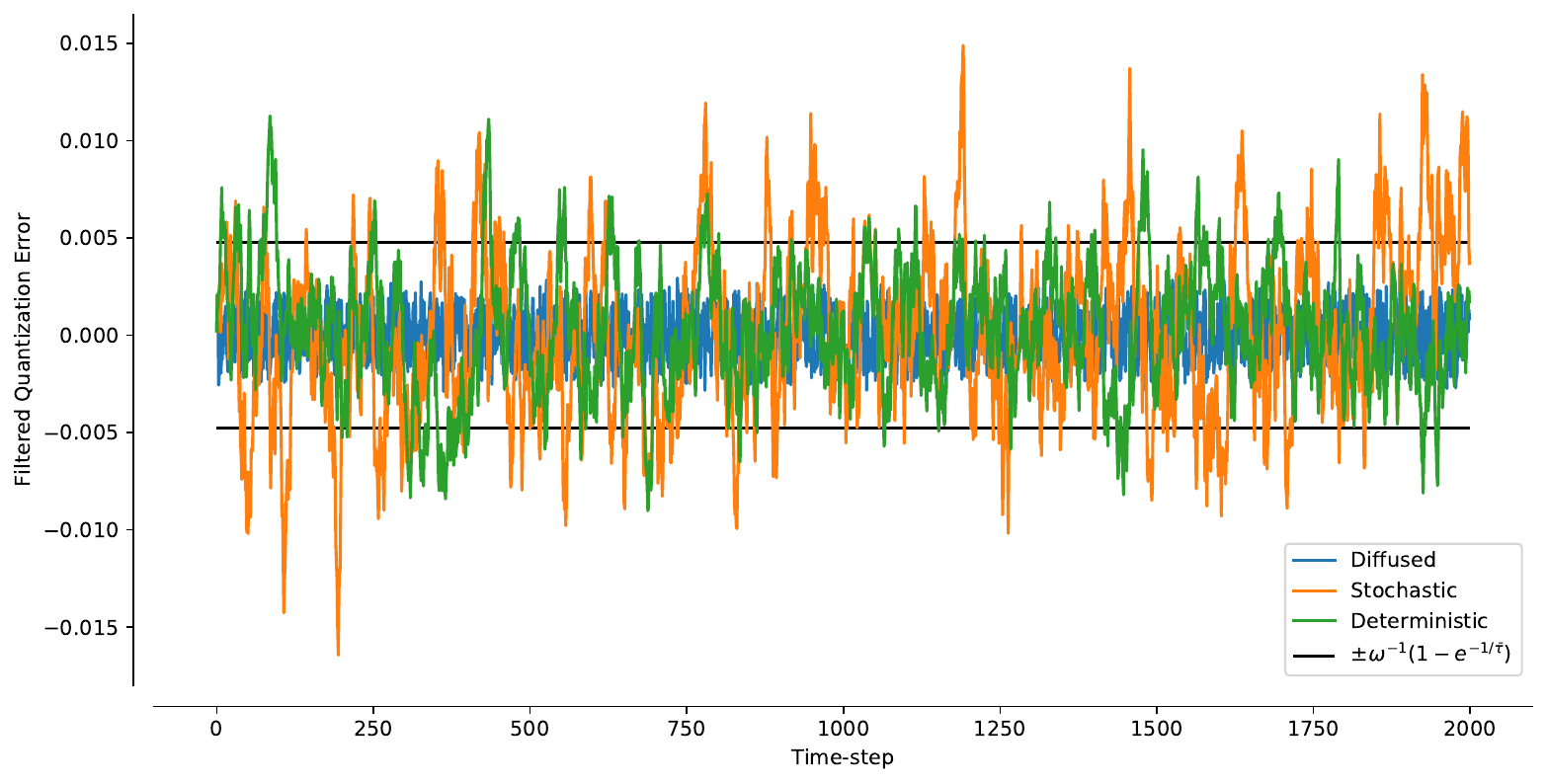}
\caption{
Numerical validation of Theorem~\ref{thm:synapse} ($\bar{\tau} = 10$) given the same input signal from Figure~\ref{fig:bounds}, and compared to the filtered error for the same two alternative quantizers.
}
\end{figure}
\end{minipage}

\section{Uniform Voltages}
\label{sec:uniform-voltage}

\begin{figure}
\centering
\begin{subfigure}{.5\textwidth}
  \centering
  \includegraphics[width=\linewidth]{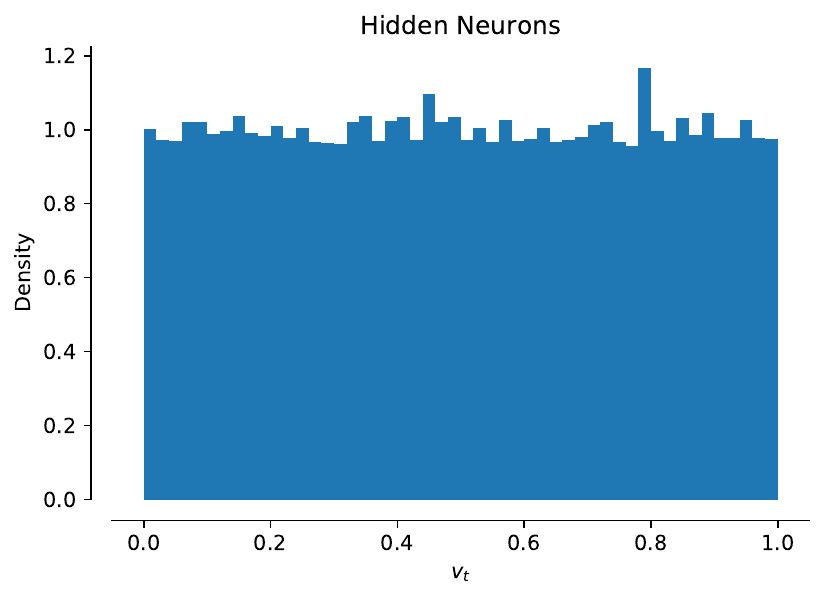}
\end{subfigure}%
\begin{subfigure}{.5\textwidth}
  \centering
  \includegraphics[width=\linewidth]{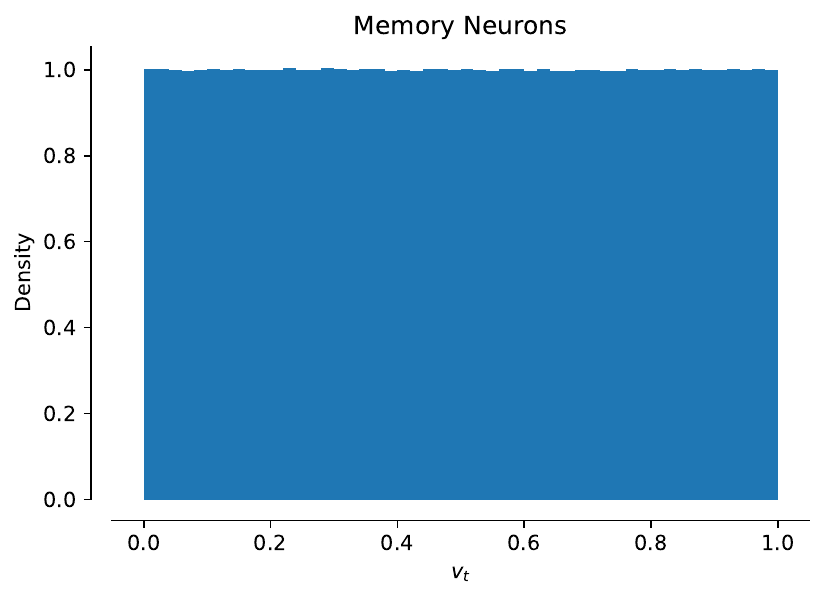}
\end{subfigure}
\caption{ \label{fig:voltages}
  Distribution of neuron state-variables ($v_t$) for the Hybrid-Spiking Legendre Memory Unit~(hsLMU) network solving the psMNIST task (see main article).
  Histograms (bin-width of $0.02$) are provided for the hidden neurons ($f_h(\cdot)$; Left) and the memory neurons ($f_m(\cdot)$; Right).
  Data is collected for all $t \in [1, 784]$ steps across the first $500$ test digits.
}
\end{figure}

Empirically, and in practice, we found that the neurons' voltages ($v_t$) tend to remain uniformly distributed over time.
Specifically, for benchmarks that we explored, $v_t$ is well-approximated as $\mathcal{U}[0, 1)$ -- especially so for the memory neurons (see Figure~\ref{fig:voltages} for example).

To quantify the assumption of uniformity, we performed two-sided Kolmogorov-Smirnov tests for goodness of fit.
The hidden neurons had an empirical CDF that was no more than 0.3\% different from that of the exact uniform distribution, while the memory neurons had an empirical CDF that was no more than 0.03\% different (KS-statistics~$< 0.003$ and $0.0003$, respectively, with $p$-values~$< 10^{-7}$).
Thus, the uniform assumption (used only by Theorem~\ref{thm:error} and its dependent claims) holds approximately for both populations.

To quantify the stronger assumption that the voltages are also independently distributed from one time-step to the next, we measured the sample covariance between $v_{t - 1}$ and $v_t$.
Although uncorrelatedness does not imply independence, we found that the memory layer had a covariance that is $\approx1\text{,}380\times$ smaller than its variance.
In contrast, the hidden layer only had a covariance that is $\approx1.25\times$ smaller than its variance.
Thus, the independence assumption (used only by Corollary~\ref{cor:gradient}) does not hold very well for the hidden layer, as its quantization errors are correlated in time.
We suspect that the temporal covariance in the hidden neurons is due to these neurons being driven by low-frequency inputs close to zero, versus the inputs to the memory neurons being sufficiently decorrelated in time.

\section{LIF Normalization}
\label{sec:lif}

For the hidden layer, $f_h(\cdot)$, we use the leaky integrate-and-fire~(LIF) neuron model with a leak of $\bar{\tau}_\text{rc} = 10$ time-steps and a refractory period of $\bar{\tau}_\text{ref} = 1$ time-step---corresponding to Nengo's defaults given a time-step of 2\,ms---such that its time-averaged response curve is:\footnote{
$\logp(x) = \ln(1 + x)$ and $\expm(x) = \exp(x) - 1$ are standard nonlinearities in NumPy and TensorFlow.
}
$$
f_h(x_t) = \frac{1}{\bar{\tau}_\text{ref} + \bar{\tau}_\text{rc} \logp \left(\frac{1}{\alpha x_t + \beta} \right)} \text{,} \quad 0 \le f_h(x_t) < 1 \text{,}
$$
where $(\alpha, \beta)$ are constants that scale and shift the curve.
Specifically, we bias the input such that $f_h(x_t) = 0 \iff x_t \le 0$, and scale it such that $f_h(1) = r \overset{\text{DEF}}{=} e / (1 + e)$.
This is achieved by choosing:
$$
\alpha = -\left[\expm \left( \frac{\bar{\tau}_\text{ref} - 1 / r}{\bar{\tau}_\text{rc}} \right)\right]^{-1} - 1 = -\left[\expm \left( \frac{\bar{\tau}_\text{ref} - (1 + e) / e}{\bar{\tau}_\text{rc}} \right)\right]^{-1} - 1 \text{,}
$$
and $\beta = 1$.
Since $(\alpha, \beta)$ are constants, they can be fused with existing parameters in the final network, in particular by multiplying the incoming weights with $\alpha$ and adding $\beta$ to the incoming bias parameters.

\begin{figure}
\centering
\begin{subfigure}{.5\textwidth}
  \centering
  \includegraphics[width=\linewidth]{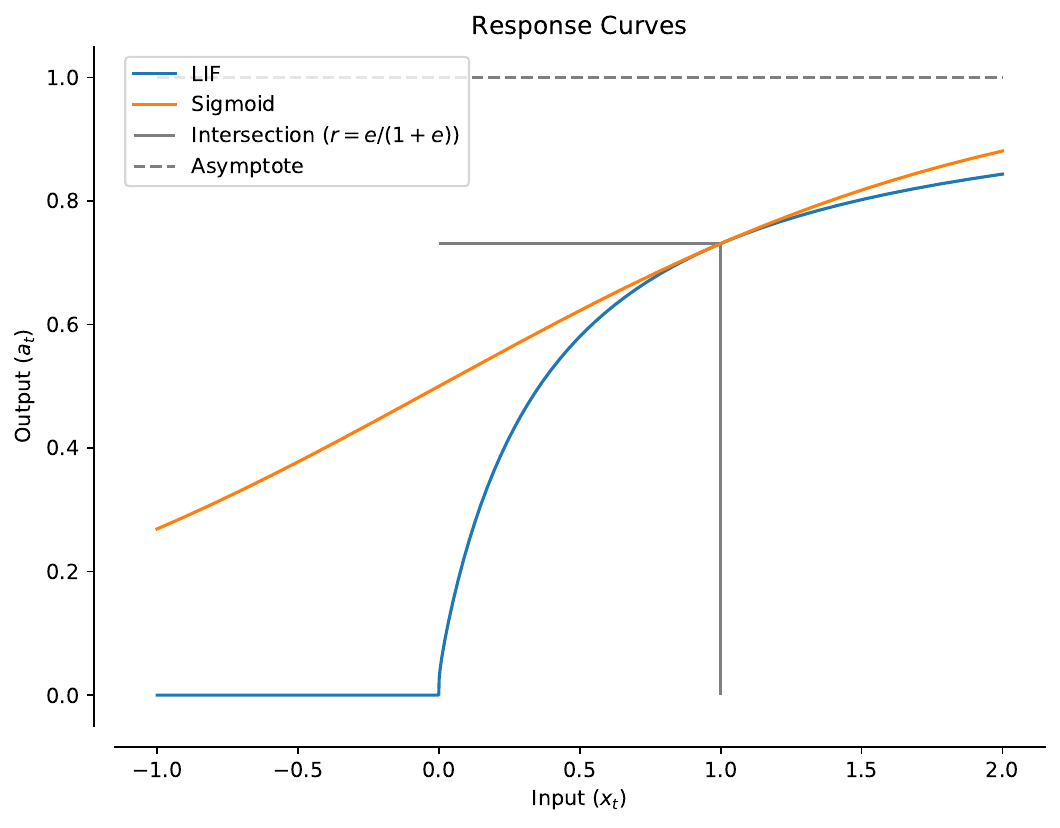}
\end{subfigure}%
\begin{subfigure}{.5\textwidth}
  \centering
  \includegraphics[width=\linewidth]{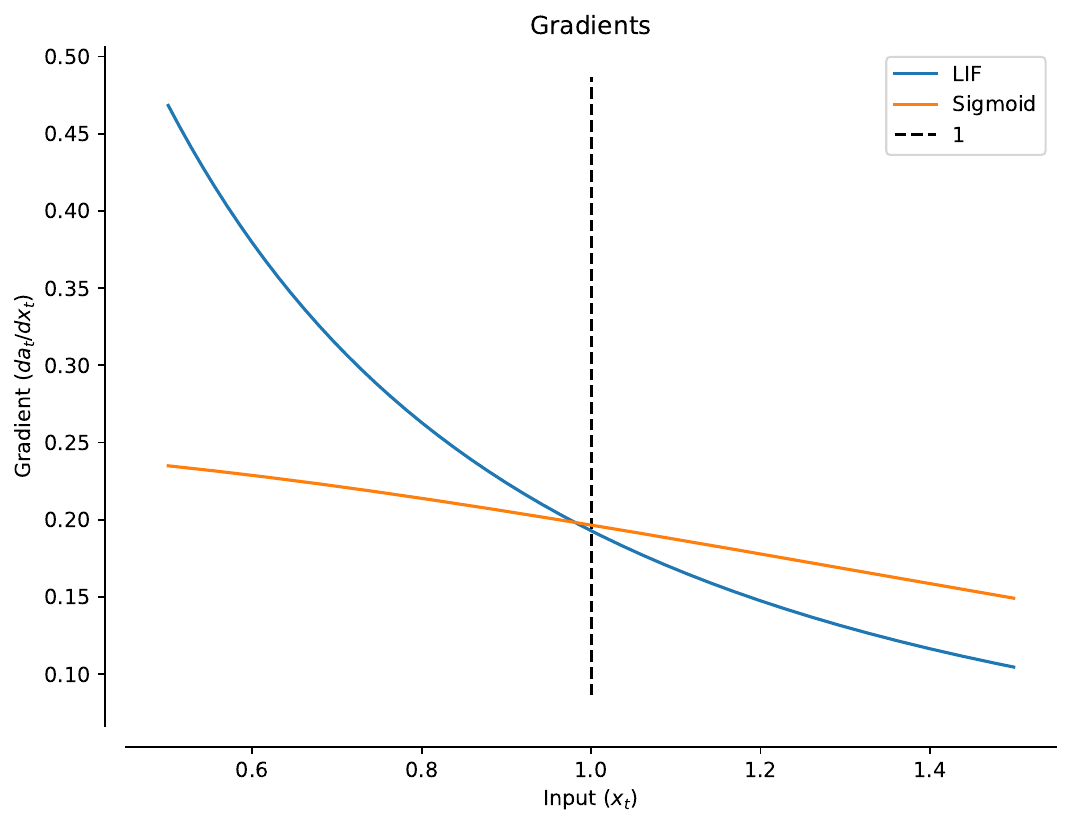}
\end{subfigure}
\caption{ \label{fig:lif}
The leaky integrate-and-fire~(LIF) with discrete time-constants $(\bar{\tau}_\text{rc}, \bar{\tau}_\text{ref}) = (10, 1)$ and a gain and bias of $(\alpha, \beta)$ such that $0 \le f_h(x_t) < 1$, $f_h(x_t) = 0 \iff x_t \le 0$, and $\omega = 1$ corresponds to a biological time-step of $2$\,ms.
(Left)~The response curve is scaled such that $f_h(1) = \sigmoid(1) = r = e / (1 + e)$. (Right)~The gradients are also approximately equal at $x_t = 1$, and become identical in the limit of $\bar{\tau}_\text{rc} \rightarrow \infty$.
}
\end{figure}

The above ``time-normalized LIF'' is convenient for two reasons.
The first reason is that the nonlinearity becomes normalized to $0 \le f_h(x_t) < 1$, such that a value of $\omega = 1$ corresponds to a spiking LIF model with a biological time-step of $2$\,ms, analogous to a conventional LIF neuron model that spikes at most once per step.
The second reason is that $f_h(1) = \sigmoid(1) = r$, and so the model has a similar dynamic range to the $\sigmoid(\cdot)$ activation (see Figure~\ref{fig:lif}), making it somewhat more compatible with weight initialization methods designed for sigmoidal layers.
We also note that the gradients of $f_h$ and $\sigmoid$ are approximately the same at $1$, and become equal as $\bar{\tau}_\text{rc}$ increases.

\end{document}